# An Ontology for Social Determinants of Education (SDoEd) based on Human-AI Collaborative Approach

Navya Martin Kollapally[1], James Geller[2], Patricia Morreale[1], Daehan Kwak[1]

[1] KEAN University
Union, NJ 07083
[2] New Jersey Institute of Technology
Newark, NJ 07102
{nmartink, pmorreal, dkwak}@ kean.edu
{james.geller}@njit.edu

**Abstract**

The use of computational ontologies is well-established in the field of Medical Informatics. The topic of Social Determinants of Health (SDoH) has also received extensive attention. Work at the intersection of ontologies and SDoH has been published. However, a standardized framework for Social Determinants of Education (SDoEd) is lacking. In this paper, we are closing the gap by introducing an SDoEd ontology for creating a precise conceptualization of the interplay between life circumstances of students and their possible educational achievements. The ontology was developed utilizing suggestions from ChatGPT-3.5-010422 and validated using peer-reviewed research articles. The first version of developed ontology was evaluated by human experts in the field of education and validated using standard ontology evaluation software. This version of the SDoEd ontology contains 231 domain concepts, 10 object properties, and 24 data properties.

# 1 Introduction

According to the US Department of Health and Human Services (HHS), Social Determinants of Health (SDoH) [1] are the conditions in the environment where people are born, live, learn, work, play, and age that affect the quality-of-life outcomes and risks. Education, health, and well-being are intrinsically interconnected. Education profoundly impacts individuals' lives, playing a crucial role in alleviating poverty and diminishing socioeconomic and political disparities. According to a study by the Centers for Disease Control and Prevention [2], high school students who demonstrated higher academic performance showed a greater tendency towards better health-related behaviors and a notably lower prevalence of health-related risk behaviors when compared to students who exhibited poor academic performance. In analogy to SDoH, these factors have variously been referred to as Social Determinants of Education (SDoEd) or simply SDE [3], e.g., lack of access to a high-speed internet connection.

Given the significant influence of education on individuals' lives and its role in addressing poverty and reducing inequalities, it is important to establish an ontology in this domain. It can serve as a comprehensive framework for organizing and representing knowledge related to education and its impact on society. By capturing the relationships, concepts, and interdependencies within the educational landscape, an ontology can facilitate better understanding, analysis, and decision-making.

Computationally, an ontology is a hierarchical structure of concepts, where pairs of concepts are connected by IS-A (generalization) links and semantic links. Concepts may also have their own local attributes. In a diagram, an ontology appears as a nodes-and-links graph. Refer to Figure 2 in the Results Section for an intuition of such a diagram. Bubbles represent concepts, and arrows are IS-A links. General concepts are at the left, and specific concepts are at the right. Notably, some of these determining factors might be circular and mutually reinforcing. For example, bad health will lead to poor school attendance, which could in turn lead to not learning about a healthy lifestyle or not being able to get into college, perpetuating social issues associated with low income and low living standards, closing the cycle by not being able to afford good healthcare.

The risk factors of SDoEd are not restricted to racial and ethnic minorities as they are often income-based, but these populations are at a higher risk compared to their white peers. The research goal of this work is to present an ontology for Social Determinants of Education (SDoEd). A human-AI collaborative approach to concept collection using ChatGPT-3.5-010422 was utilized. Along with the design and development of the ontology, the human expert and software-based evaluation criterion for ensuring consistency and coherence of the ontology are presented in this work.

## 2    Background

An existing framework for Determinants of Education/Learning involves asking nine inter-dependent questions related to education; the answers to these questions will produce the concepts for School Health Education (SHE) [2]. To create a comprehensive ontology, it is crucial to compile an extensive list of terms and concepts that cover the domain under consideration. When enriching a domain ontology, developers may rely on research articles to gather concepts that expand the ontology's scope and coverage. However, despite a thorough search, the tasks of gathering all relevant concepts and ensuring the ontology's comprehensiveness were challenging. To address this issue, we utilized a Generative Pretrained Transformer (GPT), a language model trained on extensive text datasets. OpenAI's ChatGPT [4], built on the GPT-3.5/4/4o, is a chatbot that utilizes supervised and reinforcement learning techniques to generate human-like responses to natural language prompts and was trained on licensed and publicly available data through 2023.

Rather than relying solely on the concept choices suggested by ChatGPT-3.5-010422, we ensured the validity of the concepts and their relationships by cross-referencing them with published articles from reputable sources such as PubMed Central (PMC) [5], International Journal of Education Research (IJER) [6], and American Educational Research Association (AERA) [7].

## 3 Methods

We utilized the ontology principles stated by Noy [8] for developing the ontology for SDoEd. Furthermore, we have used the design and evaluation criteria as used in [9-11]

*3.1 Domain and scope of ontology*
The scope of Social Determinants of Education (SDoEd) encompasses a wide range of factors that impact student engagement in education and substantiate the existence of an achievement gap. As defined by the American Board of Education, the achievement gap occurs when there is a statistically significant disparity in average scores between different groups of students [12], typically categorized by race/ethnicity or gender. These factors can be influenced by a variety of elements, including political, economic, cultural, and societal factors.

*3.2 Enumerate important concepts for developing SDoEd*
We retrieved articles and reports available from trustworthy sources by performing a keyword-based search on the web. These sources utilized terms such as "social determinants of education," "role of education in SDoH," "educational disparities," "reasons for achievement gap in education," and "determinants of learning." By analyzing the results, the main categories of SDoEd were identified to achieve comprehensive coverage of relevant domain concepts.

To confirm coverage in terms of concepts and to address any potential gaps, we then utilized ChatGPT-3.5-010422. We used specific prompts such as "Main categories of Social Determinants of Education," "Sub-concepts related to Economic stability that contribute to the achievement gap," "Child concepts associated with Parental factors influencing educational determinants," "Is there an IS-A relationship between factors affecting health and well-being and the parent concept of Social Determinants of Education," and "Does cyberbullying fall under the child concept of technology integration?" These prompts helped us to find more candidate concepts and to clarify the relationships and classifications within the broader context of SDoEd.

Before adding each of the concepts under a main category of the ontology, we validated the IS-A relationships by searching for articles in PMC within the range of 2018-2023, in the IJER, and on the websites of the AERA and the Department of Education. We utilized the advanced query feature of PMC to validate the concept pairs suggested by ChatGPT-3.5-010422.

During the validation of concept pairs from ChatGPT-3.5-010422 in the relevant sources, we encountered new concepts that were not in the output lists from ChatGPT-3.5-010422. Hence in addition to *forward validation,* i.e., validating concept pairs extracted from ChatGPT-3.5-010422 utilizing the target sources, we also performed *backward validation*, i.e., extracting concept pairs from target sources and validating them using ChatGPT-3.5-010422. Figure 1 represents the forward validation in which ChatGPT-3.5-010422 states that "availability of after-school programs" is a child concept of "availability of educational resources." To validate this concept pair, we used the prompt "how availability of after-school program and educational resources affect social determinants of education" in PMC, IJER, the AERA, and the Department of Education websites. After identifying relevant articles, a concept pair (parent-child concept pair) is either accepted into the SDoEd ontology or rejected. For backward validation while performing a manual review of relevant articles from target sources, new concept pairs may be identified. These concept pairs will be framed as two concepts connected by an IS-A relationship as shown in Figure 1. We prompted ChatGPT with text corresponding to "Does this RDF triple share a valid IS-A relationship?" (RDF is the Semantic Web Resource Description Framework.)

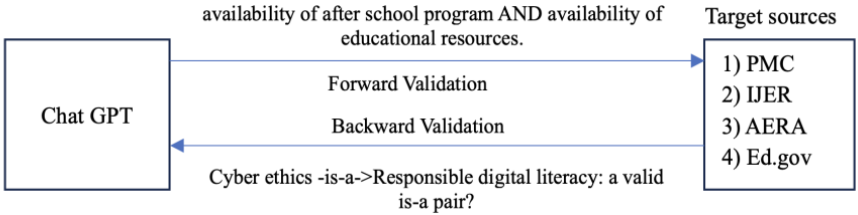

Figure 1: Visualization of forward and backward validation.

### 3.3 Concept Categorization
After the initial step of concept extraction from ChatGPT-3.5-010422, scholarly articles, and government educational websites, we placed the concepts under six main categories. They are:

    a) Cultural factors: This parent concept includes child concepts that significantly shape the educational environment and practices within a particular community or society [13]. They can influence how education is valued, the expectations placed on students, the teaching and learning methods employed, and the overall educational goals and priorities.

    b) Economic factors affecting education: This category includes concepts referring to the financial resources and socio-economic conditions that play a

significant role in shaping educational opportunities and outcomes [14]. These factors encompass aspects such as funding and resource allocation, socioeconomic disparities, and access to educational resources.

c) Factors influencing health and well-being: The sub-concepts under this main concept encompass a range of elements that impact the physical, mental, and emotional well-being of individuals, which in turn can affect their educational experiences and outcomes.

d) Institutional factors influencing education: Institutional factors encompass the policies, structures, and organizations within the education system that directly or indirectly impact educational outcomes [15].

e) Neighborhood factors influencing education: Neighborhood factors comprise the characteristics and conditions of the local community that surrounds a school, which can significantly influence educational opportunities [16].

f) Parental factors: Parental factors refer to the influences, actions, and characteristics of parents or guardians that significantly impact educational opportunities and outcomes for children [17]. Parental factors play a crucial role in shaping children's educational experiences, motivation, and academic achievements, as parents serve as primary caregivers and key influencers in their children's educational journey.

## 3.4 Developing an SDoEd Ontology

To implement the SDoEd ontology, we utilized Protégé 5.5, an open-source ontology editor by Stanford University [18]. The SDoEd ontology was developed as a Web Ontology Language (OWL) file. Protégé refers to "concepts" as "classes," and allows adding properties (~attributes) and relationships between the classes. The class "Thing" is predefined in Protégé and is used as the root of every ontology created with it. Protégé enables users to edit ontologies in OWL and use a HermiT reasoner to validate the consistency and coherence of the developed ontologies.

## 3.5 Software-based SDoEd Ontology Evaluation

We performed consistency checking in Protégé by utilizing HermiT [19] Version 1.4.3.456. The HermiT reasoner is based on hyper tableau calculus, which allows it to avoid nondeterministic behaviour exhibited by the tableau calculus that is utilized in FaCT++ [20] and Pellet [21]. Nondeterministic behaviour arises when tableau calculus may have to make arbitrary choices that can lead to inefficiency, and this is avoided by structuring the reasoner process as in hyper tableau calculus.

## 3.6 Human Expert SDoEd Ontology Evaluation

The main goal of the evaluation of an ontology is to make sure that it is consistent, accurate, and maintains a high level of adaptability and clarity. After evaluating the SDoEd ontology with HermiT for consistency and coherence, we involved two human expert evaluators with extensive experience in the field of education. To understand the percentage agreement between the two evaluators, we utilized Cohen's kappa coefficient (κ). κ is an alternative when the overall accuracy is biased to understand the level of agreement between two evaluators. Both human evaluators (P1 and P2) were provided with the same spreadsheet of 100 randomly selected concept pairs. Among the 100 concept pairs, we provided 10 concept pairs as training samples to present the flavor of the ontology and 90 concept pairs that needed to be evaluated. The spreadsheet contained three kinds of concept pairs: pairs related by IS-A, pairs related as ancestor/grandparent-child, and pairs that were not hierarchically related. P1 and P2 were aware of the fact that the spreadsheet contained these three kinds of concept pairs. Table 1 provides examples of the concept pairs included in the spreadsheet.

Table 1: A snippet of the concept pairs provided to the human expert.

| Parent | Relationship | Child | Related | Farther away | Reason if unrelated |
|---|---|---|---|---|---|
| Availability of educational resources | <--is--a | Availability of tutoring centers | Yes | | |
| Neighborhood safety | <--is--a | Emotional and Social Development in early years | Yes | | |
| Sleep quality of children | <--is--a | Discipline polices | Yes | | |
| Parental factors | <--is--a | Parental educational level | Yes | | |
| Sleep quality of children | <--is--a | Technology integration policies | | Yes | Sleep quality of children is unrelated to technology integration |

The 10 samples provided to the evaluators included five of the 'Child' fields filled with 'No' and corresponding reasons were provided in 'Reason if unrelated,' three of the 'Child' fields filled with 'Yes,' and two of the 'Farther away' fields with Yes. For each pair, the fourth column ('Child?' in Table 1) had to be filled in with 'Yes,' if the evaluator felt that the concepts were connected by a parent-child (IS-A) relationship, and 'No,' otherwise. If the answer was 'No,' they were asked to fill in the reason in the column 'Reason if unrelated.' These reasons provided us with directions on how to make improvements to the design of the ontology. The evaluators were asked to fill in the 'Farther away' column with 'Yes,' whenever they felt that the concepts were related by a grandparent or ancestor relationship, i.e., a chain of IS-As. The evaluators were also asked to give reasons in this case. P1 and P2 independently reviewed the pairs, and we used an online κ calculator [22] to identify the level of agreement.

A κ > 0.4 is considered as moderate agreement, κ > 0.6 indicates substantial agreement, and κ = 1 means perfect agreement.

# 4 Results

We could not locate any preexisting domain ontology specific to SDoEd. This supports the need for our research work. We also used Protégé for evaluation (HermiT). The class metrics returned by Protégé/HermiT are in Table 2.

Table 2: Class metrics from Protégé.

| Metrics | Count |
|---|---|
| Axioms | 498 |
| Logical axiom count | 267 |
| Declaration axiom count | 231 |
| Class count (not counting Thing) | 231 |

Our SDoEd ontology, developed in Protégé, is available as an OWL file in GitHub [23] . In Figure 2 is a snippet of the ontology, visualized using the OWLViz plugin of Protégé. The SDoEd ontology is coherent and consistent as per the HermiT reasoner, also available as a plug-in in Protégé. The confusion matrix for evaluators is given in Table 3, and a κ value of 0.6345 was obtained in the first round of evaluation. This indicates a substantial agreement (83.389%), hence no mitigation plan and no second round were necessary. The κ value represented that the experts were in "*substantial agreement"* about the domain coverage of the designed ontology. For concept extraction and backward validation, a total of 72 prompts were posed to ChatGPT-3.5-010422. These prompts encompassed a wide range of topics and concepts to ensure comprehensive coverage. The extracted concepts were then validated to ensure accuracy and relevance in the given context. Figure 3 shows the six main categories and few of direct subcategories of the SDoEd ontology, full view in the GitHub repository [23]. Even though details are hard to see in the figure, it provides an overall "Gestalt" of the ontology.

Table 3: Confusion matrix of evaluator 1 and evaluator 2.

| Confusion matrix | Hierarchical related concept pairs | Unrelated concept pairs |
|---|---|---|
| Evaluated as hierarchical related concept pairs by evaluator 1 | 46 | 9 |
| Evaluated as hierarchical un-related concept pairs evaluator 1 | 19 | 16 |
| Evaluated as hierarchical related concept pairs by evaluator 2 | 52 | 0 |
| Evaluated as hierarchical un-related concept pairs by evaluator 2 | 21 | 17 |

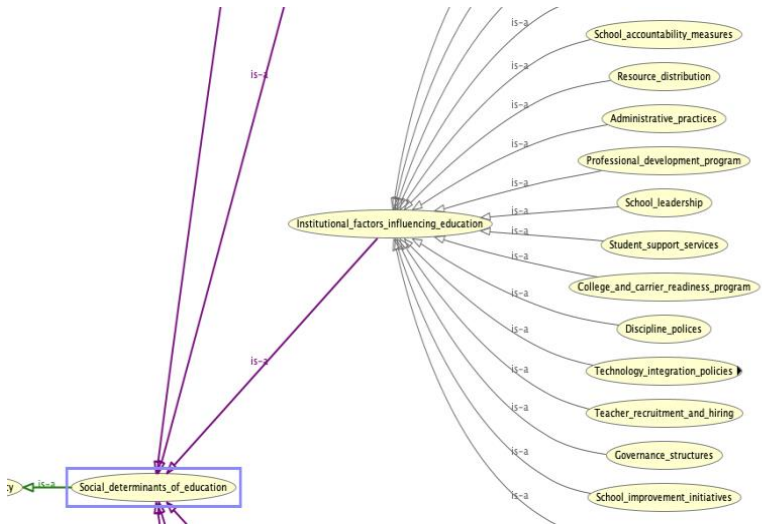

Figure 2: Snippet from OWLViz visualization of SDoEd ontology.

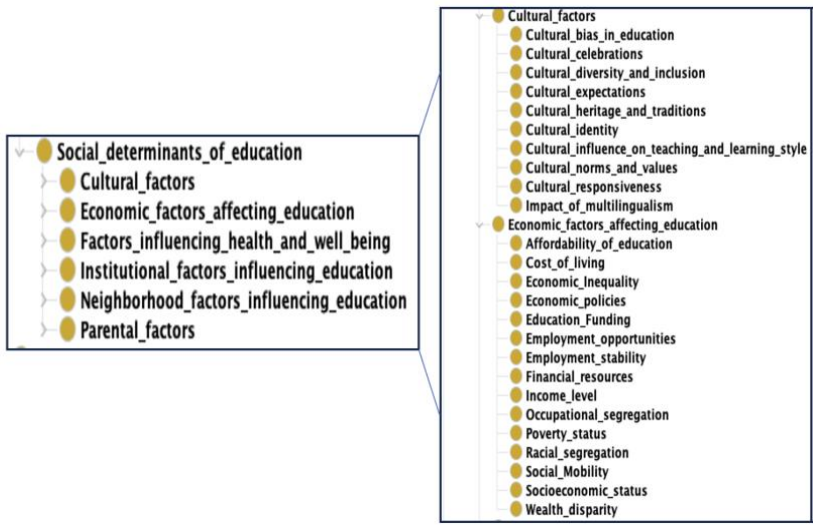

Figure 3: Six main categories and few of direct subcategories of SDoEd ontology.

# 5 Conclusions

The ontology for Social Determinants of Education holds significant potential in enhancing our understanding of the complex interplay between education and various socio-environmental factors, including health. By creating a comprehensive framework that captures the concepts, relationships, and dependencies within this domain, the ontology can serve as a tool for organizing and representing knowledge related to educational disparities, poverty alleviation, and reducing inequalities. This research contributes to the broader goal of leveraging data-driven and intelligent systems to enhance educational outcomes and promote equity considering pressing challenges, such as the recognition of structural racism. This prototype of the SDoEd ontology contains 231 concepts, 10 object properties, and 24 data properties. It is available in the GitHub repository [23].

# 6  Limitations and Future Work

To facilitate the utilization of the Social Determinants of Education ontology in natural language processing (NLP) tasks, we plan to annotate the ontology using CURIES IDs, which are shortened, standardized references that simplify concept identification. This will enhance the accessibility and interoperability of the ontology. Additionally, to enhance the richness of relationships within the SDoEd ontology, more contextual information will be incorporated. The authors also plan to explore various prompt engineering techniques and train an LLM for supporting ontology development.

## Acknowledgments

During the final editing of several sections of this paper, ChatGPT was used to check for grammar and expression errors. No changes of the content were performed by it.